\documentclass{article}
\date{\vspace{-5ex}}

\pdfoutput=1

\usepackage[final]{corl_2020} 
\usepackage{amsmath} 

\title{DeepMPCVS: Deep Model Predictive Control for Visual Servoing}

\usepackage{titling}
\usepackage{graphicx}
\usepackage[export]{adjustbox}
\usepackage{setspace}
\usepackage{subfigure}
\usepackage{mathptmx}
\usepackage{bm}
\usepackage{amsmath,amssymb} 
\usepackage{color}
\usepackage{algorithm}
\usepackage{algpseudocode}
\usepackage{enumerate}
\usepackage{epstopdf}
\usepackage{array}  
\usepackage{eqparbox}
\usepackage{multirow}
\usepackage{url}
\usepackage{mathtools}
\usepackage{hyperref}
\usepackage{adjustbox}
\usepackage{caption}
\usepackage{textcomp}
\graphicspath{ {./figures} }
\usepackage{xcolor}
\usepackage{wrapfig}
\usepackage{authblk}

\usepackage[T1]{fontenc}
\usepackage[english]{babel}
\usepackage{algorithm,algorithmicx,algpseudocode}
\usepackage{amssymb}
\usepackage{mathtools}

\usepackage{caption} 
\usepackage{multicol}

\usepackage{array}
\newcolumntype{P}[1]{>{\centering\arraybackslash}p{#1}}

\author[1,*]{\textbf{Pushkal Katara}}
\author[1,*]{\textbf{Y V S Harish}}
\author[2]{\textbf{Harit Pandya}}
\author[1]{\textbf{Abhinav Gupta}}
\author[1]{\textbf{AadilMehdi Sanchawala}}
\author[3]{\mbox{\textbf{Gourav Kumar}}}
\author[3]{\textbf{Brojeshwar Bhowmick}}
\author[1]{\textbf{K. Madhava Krishna}}
\affil[1]{Robotics Research Center, IIIT Hyderabad, India}
\affil[2]{University Of Lincoln, UK}
\affil[3]{TCS Innovation Labs, Kolkata, India}

\begin{document}

\maketitle

{\let\thefootnote\relax{\footnote{$*$: denotes equal contribution}}}



\begin{abstract}
The simplicity of the visual servoing approach makes it an attractive option for tasks dealing with vision-based control of robots in many real-world applications. However, attaining precise alignment for unseen environments pose a challenge to existing visual servoing approaches. While classical approaches assume a perfect world, the recent data-driven approaches face issues when generalizing to novel environments. In this paper, we aim to combine the best of both worlds. We present a deep model predictive visual servoing framework that can achieve precise alignment with optimal trajectories and can generalize to novel environments. Our framework consists of a deep network for optical flow predictions, which are used along with a predictive model to forecast future optical flow. For generating an optimal set of velocities we present a control network that can be trained on-the-fly without any supervision. Through extensive simulations on photo-realistic indoor settings of the popular Habitat framework, we show significant performance gain due to the proposed formulation vis-a-vis recent state of the art methods. Specifically, we show a faster convergence and an improved performance in trajectory length over recent approaches.

\end{abstract}



\section{Introduction}
The concept of visual servoing has been widely used in the field of robotics for a variety of tasks from manipulation to navigation \cite{vs_nav1,vs_nav2,vs_nav3,vs_manip1,vs_manip2}. Mapping the observation space to the control/action space is the fundamental objective behind most of the visual servoing problems. Classically, visual servoing is highly dependent on handcrafted image features, environment modeling, and an accurate understanding of system dynamics for the generation of control commands \cite{vsbasic}. 
Sensitivity to the accuracy of the available information also prevents such systems from adjusting to the statistical irregularities of the world. 
To mitigate these issues, recent approaches aim to learn supervised models that predict relative camera poses between the current and the desired images \cite{deepvs1,deepvs2, deepvs3}. Although this overcomes the requirements of having handcrafted image features or any explicit information about the environment, it fails to generalize in unseen environments. Other approaches further attempt to exploit information from intermediate representations like optical flow \cite{DFVS}, however, they use classical visual servoing controller which only looks one-step ahead hence results in sub-optimal trajectories. On the other hand, many deep reinforcement learning based approaches such as \cite{targetdriven} and \citep{splitnet} do not show their convergence in a 6-DoF domain due to an intractable number of samples in higher dimensional actions and continuous state space. This narrows down their scope of performance in real-world visual servoing tasks for precise alignment. Furthermore, they also need to be extensively retrained in newer environments as a result, their generalisation capabilities are limited. A few approaches \cite{deepforesight,dvmpc}, employ a model based learning framework for visual servoing, they attempt to learn a forward predictive model to represent the process and a policy for planning over the predictive model. However, their predictive model is trained to minimise the appearance loss which is difficult to optimise for a longer horizon and does not generalize to novel environments. Furthermore, the policy is obtained either through sampling \cite{deepforesight} or by learning from human demonstrations \cite{dvmpc} that do not scale to a higher dimensional action space.

In this work, we present a deep model predictive control strategy that exploits the visual servoing concept in a principled fashion. This is achieved by formulating a novel state prediction model for visual servoing based on dense off-the-shelf/unsupervised optical flow predictions. The prediction model is employed to generate optimal velocity commands using a control network, we further propose an efficient unsupervised strategy for online optimisation of control network. We validate our approach on a photo-realistic visual servoing dataset proposed in \cite{DFVS}, comparing against an exhaustive set of baselines among classical, deep supervised and model predictive visual servoing approaches. 



Our contributions are summarized as follows:
\begin{itemize}
    
    
    \item We propose a novel unsupervised Deep Model Predictive Control pipeline for visual servoing using optical flow as intermediate representation in 6-DoF.
    
    
    \item The control actions are learned unsupervised as the LSTM decoder strives to predict future controls over a time horizon such that the optical flow accrued as a consequence of these predictions matches the desired flow. The desired flow between the current and desired images is computed either through a supervised \cite{flownet-2} or unsupervised \cite{ddflow} framework.
    
    \item The online optimization over the MPC framework enables the proposed approach to adapt to any environment even with inaccurate flow predictions and system dynamics information. 
    
    

    \item We compare with existing deep visual servoing methods by evaluating on parameters like translation error, rotation error, trajectory length, total no. of iterations and time per iteration. Our approach exhibits significant convergence and is faster than the state of the art deep visual servoing methods \citep{DFVS}, \cite{cem}, \cite{servonet},\cite{photometricvs}. We present both qualitative and quantitative results for the same. We make the code \footnote{\href{https://github.com/pushkalkatara/DeepMPCVS}{Github Link: https://github.com/pushkalkatara/DeepMPCVS}} publicly available.

    
    
    
    
    

\end{itemize}
\section{Related Work}
\label{sec:Related_Work}

\textbf{Visual Servoing.} Visual Servo (VS) relates to the problem of attaining a goal pose in the environment using image measurements from a camera sensor attached to a robot. It can be broadly classified in two approaches: (i) PBVS: Pose based Visual Servoing where hand-crafted features from images are used to estimate geometrical information such as pose of the target, camera parameters and controller is responsible to minimize pose error between current image and goal image. (ii) IBVS: Image-based Visual Servoing approaches perform feature extraction and minimize the feature error explicitly in image space. The control law is to perform gradient descent in feature space, which is then mapped to the robot's velocities using an Image Jacobian \cite{vsbasic}. Classical visual servoing approaches \cite{ vs_nav1, vs_manip1,vs_manip2,vs_contours} employ local appearance based features such as keypoints, lines and contours to describe the scene. On the other hand, direct visual servoing approaches \cite{photometricvs, histogramvs} skip the feature extraction and directly minimize the difference in images. Although these direct approaches mitigate the issue of incorrect matches, they do not converge for larger camera transformations. One of the limitations of classical visual servoing methods is the requirement of the depth of the visual features while computing the Image Jacobian, which is generally not available in the case of a monocular camera. 


\textbf{Deep Visual Servoing.} Recent deep learning driven approaches \cite{deepvs1,deepvs2,servonet} learn deep neural networks that directly aim to predict the relative camera pose between the current and the desired image. The controller then takes a small step to minimize this error. These neural networks are learned in a supervised fashion and consequently do not generalize well to unseen environments. A very recent approach \citep{DFVS} combines deep learning with visual servoing by learning intermediate flow representations and minimizing their difference using a classical visual servoing controller. This approach effectively generalizes to novel environments in 6-DoF. However, since the controller optimizes only for a single time step, it generates sub-optimal trajectories and could get stuck in local minima as a result of this greedy behavior.

\textbf{Visual Navigation using Reinforcement Learning.} Visual servoing could also be posed as a target driven visual navigation problem. Motivated from the success attained by deep reinforcement learning (DRL) approaches in maze navigation tasks, there has been an increase in interest in applying imitation learning \cite{zeroshot} or reinforcement learning methods for visual navigation in indoor environments \cite{targetdriven, splitnet, intentionnet, subgoals,  midlevel}. In contrast to classical visual servoing approaches that greedily plan one time step, DRL approaches plan an optimal policy over a longer horizon even when there is no overlap between the current and desired image. However, such model free end-to-end learning based approaches are sample inefficient, thus do not scale to higher dimensional continuous actions and face difficulty while generalising to novel environments.

\textbf{Model Based Visual Control.}
Model predictive control (MPC) approaches have been successful in learning complex skills in robotics such as controlling quadrotors \cite{mpc_drone}, and humanoids \cite{humanoid_mpc} using accurate system dynamics. Keypoints based MPC models proposed by \cite{IBVS_MPC_1,IBVS_MPC_2} have been previously used in IBVS to generate optimal policies under the assumption of accurate matching and using a handful of keypoints. Classical MPC approaches have two limitations: firstly, they require accurate dynamics, secondly, they do not scale well in large state spaces. Recent deep learning approaches aim to improve upon this limitation by simultaneously learning the model dynamics along with the policies \cite{DeepMPC_1,DeepMPC_2}. For learning visio-motor control, recent approaches propose a deep network to learn forward dynamics by predicting future observations \cite{deepforesight,dvmpc}. To learn the policies along with dynamics, there are a few choices: (i) projecting the latent space into a Cartesian world and classical approaches to generate control \cite{optimalcontrol, vmpc_drone}. However, directly learning camera pose could be inaccurate for unseen environments and 6-DoF actions. (ii) Hirose et al. \cite{dvmpc} use a neural network to approximate policy which is learned in a supervised fashion from human  demonstrations. However, the issue with this method is obtaining a large and diverse dataset of human demonstrations which is expensive. Another approach \cite{deepforesight} uses visual MPC where model is trained using collected data, further perform MPC optimisation using CEM which combined makes the approach supervised and sampling-dependent making it less generalizable and less efficient than the proposed method. 


This work bridges the gap between classical and learning-based control for 6-DoF image-based visual servoing (IBVS) in novel environments with continuous action and state space. We select optical flow to represent our states instead of directly working with images. The dense optical flow is predicted using deep neural networks. Subsequently, we reformulate a predictive model for forecasting the evolution of states given a sequence of actions (velocities). We then learn a recurrent control network on-the-fly in an unsupervised fashion for computing an optimal set of velocity for a given goal state. We show superior performance vis-a-vis Deep Visual Servoing methods due to a receding horizon controller even as the framework generalizes to new environments without needing to retrain or finetune. The controller regresses to a continuous space of outputs over 6-DoF.

\section{Approach}
\begin{figure*}[t!]
    \begin{center}
    \includegraphics[width=12cm,height=6.5cm] {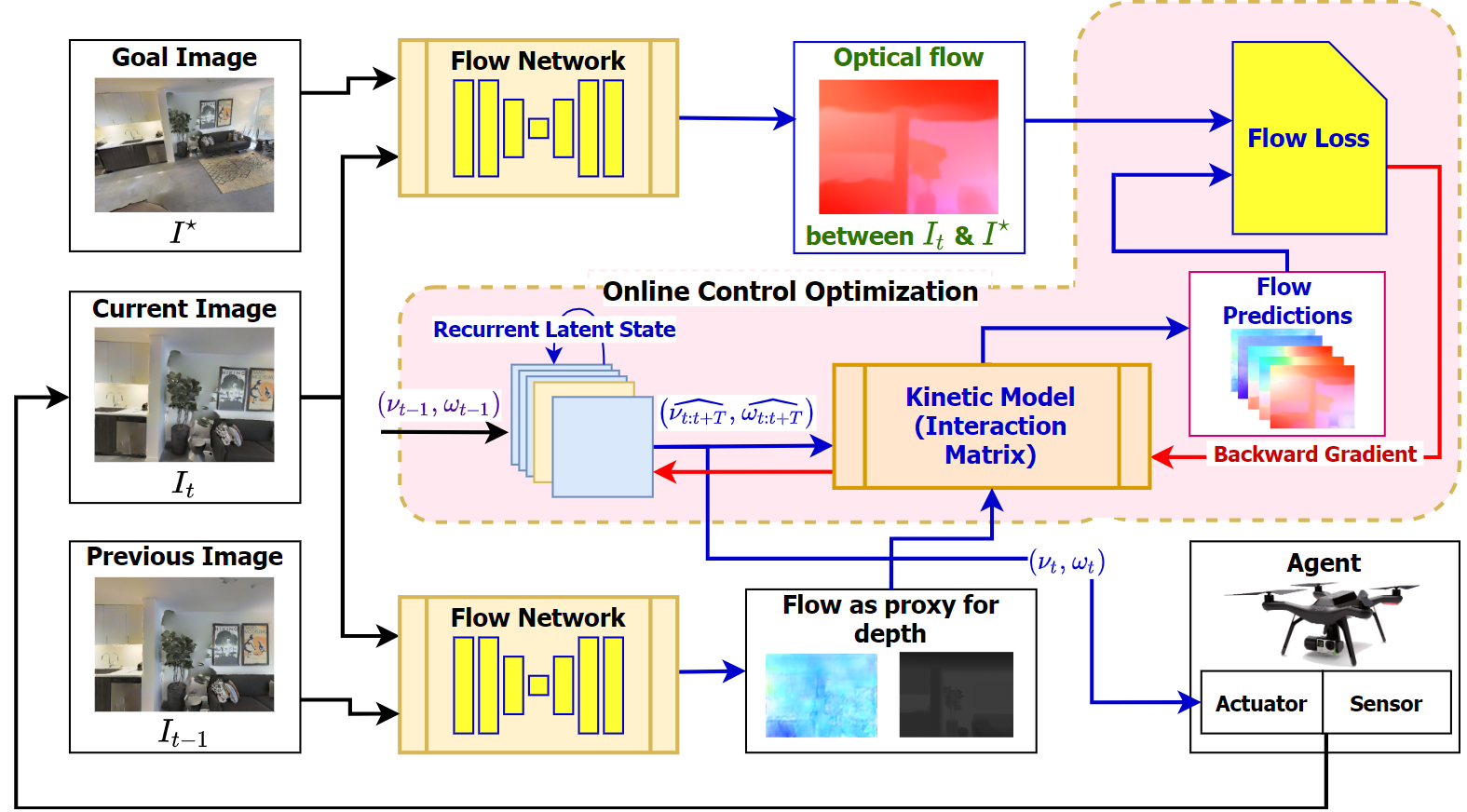}
    \caption{
    Given agent observation $I_{t}$ and desired observation $I^{*}$ the pipeline generates optimal control commands $[v_t, \omega_t]$ for each time-step t until convergence is achieved. The flow network encodes optical flow between $I_{t}$ and $I^{*}$ and the control-optimizer architecture learns to generate optimal control for each time-step $t$ in real-time through encoded flow.   
    }
    \label{fig:pipeline}
    \end{center}
    \vspace{-0.2cm}
\end{figure*}

In this work, we consider the problem of visual servoing as a target driven navigation in unseen environments using a monocular camera. Given  $I_t$, the observation of the robot in the form of monocular RGB image at any time instant $t$ and the desired observation $I^*$, our goal is to generate optimal control commands $[v_t, \omega_t]$ in 6-DoF that minimizes the photometric error between $I^*$ and $I_t$. We assume that the camera is attached to the robot (also known as eye-in-hand configuration) and is calibrated. We further assume that the environment is collision free and there exists a partial overlap between the initial and desired image. In contrast to existing visual navigation approaches \cite{dvmpc}, we aim to generate trajectories in continuous actions (in 6-DoF) and state space. Figure \ref{fig:pipeline} provides an overview of our approach. 

\subsection{Learning Intermediate Representation}
Instead of directly planning in image observation space, we employ optical flow as an intermediate visual representation for encoding differences in images. Optical flow encodes displacement for every pixel which is more relevant information for feature tracking and matching as compared to pixel intensities. As a result, optical flow has been successfully used in motion estimation \cite{ldof} and visual servoing \cite{flowcontrol,DFVS}. Furthermore, in the presence of translation camera motion in a static environment, dense flow features could also be used for estimating image depth \cite{flowdepth}. Although we do not constraint ourselves to any specific network for flow or depth estimation, in this work we use a pre-trained neural network,  Flownet 2 \cite{flownet-2} for flow estimation without any fine-tuning. To showcase that our approach can achieve similar performance even when trained without any supervision, we replace Flownet 2 with another network architecture DDFlow \cite{ddflow}, which is trained from scratch using only images in an unsupervised manner. For specifics regarding training of DDFlow, refer to the supplementary material. 


\subsection{Predictive model for visual servoing}
\label{subsec:Deep_mpc_vs}
The fundamental objective of IBVS is to minimize the error in visual features between current and desired images, $e_t = \Vert s(I_t) - s(I^*)\Vert$, which is solved by performing gradient descent in feature space, resulting in following control law:
\begin{equation}
\label{eq:classical_IBVS}
  \mathbf{v}_t = -\lambda L^+(s(I_t)-s(I^*))
\end{equation}

where, $\mathbf{v}_t = [v_t, \omega_t,]$ is the velocity command generated by the controller in 6-DoF, $s(I_t)$ and $s(I^*)$ are the visual features extracted from images $I_t$ and $I^*$ respectively. The gradient descent step size is depicted by $\lambda$ and $(.)^+$ denotes psuedo-inverse operation. The image Jacobian or the interaction matrix mapping the camera velocity to rate of change of features is given by, 
\begin{equation}
\label{eq:interaction}
L(Z_t) = \begin{bmatrix}
    -1/Z_t & 0 & x/Z_t & xy  & -(1+{x}^2) & y \\
    0 & -1/Z_t & y/Z_t & 1+{y}^2  & -xy & -x
    \end{bmatrix}.
\end{equation}
Where, $x$ and $y$ are the normalized image coordinate and $Z_t$ is the scene's depth at $(x,y,t)$. For a small time interval $\delta t$, IBVS control law, i.e. equation (\ref{eq:classical_IBVS})  can be reformulated as $\textstyle s_{t + \delta t}= s_t+L(z)\mathbf{v}_t$, which can be rewritten using optical flow as visual features as,
\begin{equation}
\label{eq:predictive_model}
\textstyle \mathcal{F}(I_t, I_{t+\delta t})= L(Z_t)\mathbf{v}_t.
\end{equation}
This equation gives us the predictive model that could be employed to reconstruct the future states (flows in our case) from the current velocity. This also points out that generating an optimal set of actions in terms of velocities $\mathbf{v}^*_{t+1:t+T}$ is a critical step for the convergence of the system towards the desired image. We achieve this task by online trajectory optimization using the predictive model and a control network described in equation (\ref{eq:predictive_model}) as described in section \ref{subsec:online_trajectory_opt}.

\subsection{Online Trajectory Optimization using Neural MPC}
\label{subsec:online_trajectory_opt}
Model predictive controllers are proven to work well for a wide variety of tasks in which accurate analytical models are difficult to obtain \cite{DeepMPC_1}. A model predictive controller aims to generate a set of optimal control actions $\mathbf{v}^*_{t+1:t+T}$ which minimize a given cost function $C(\hat{X}_{t+1:t+T},\mathbf{v}_{t+1:t+T})$ over predicted state $\hat{X}$ and control inputs $\mathbf{v}$ for some finite time horizon $T$.
In order to formulate visual servoing as MPC with intermediate flow representations, we define our cost function as mean squared error in flow between any two given images. Then MPC objective, 
\begin{equation}
\label{eq:Our_cost_function}
\textstyle \mathbf{v}^*_{t+1:t+T}=  \underset{\mathbf{v}_{t+1:t+T}}{\arg\min}  
 \lVert \mathcal{{F}}(I_t, I^*)- \widehat{\mathcal{F}}(\mathbf{v}_{t+1:t+T}) \rVert  
 \end{equation}
 is then to generate a set of velocities $\textstyle \mathbf{v}^*_{t+1:t+T}$ that minimise the error between the desired flow $\mathcal{{F}}(I_t, I^*_t)$ predicted by the flow network and the generated flow $\widehat{\mathcal{F}}(\mathbf{v}_{t+1:t+T})$. Exploiting the additive nature of optical flows (i.e. $\mathcal{{F}}(I_1,I_3) = \mathcal{{F}}(I_1,I_2)+\mathcal{{F}}(I_2,I_3)$), the generated flow could be written as:  
 \begin{equation}
 \begin{split}
 \widehat{\mathcal{F}}(\mathbf{v}_{t+1:t+T}) 
 = \sum_{k=1}^{T}[L(Z_{t})\mathbf{v}_{t+k}]
 \end{split}
\end{equation}
 using the predictive model from equation (\ref{eq:predictive_model}). This formulation provides optimal trajectory over a horizon $T$ as compared to classical IBVS controller, equation (\ref{eq:classical_IBVS}) that greedily optimises a single step. Another advantage of using the proposed approach is significant decrease in computation time as compared to classical IBVS controller,  as we avoid the overhead of matrix inversion required for computation of pseudo inverse $L^+=(L^TL+\mu diag(L^TL))^{-1}$.

The above formulation allows us to control in continuous 6-DoF action space. However, since the dimensionality of our states (flow representations) are large, classical MPC solvers could not be used to optimise equation (\ref{eq:Our_cost_function}). To tackle this issue, Finn and Levine \cite{deepforesight} use a sampling based Cross Entropy Method (CEM), but the approach is model-based sampling which requires large number of iterations to train and is less generalizable as supervised on a dataset. Another recent approach  \cite{dvmpc} employs a neural network for policy generation. They train their network using human demonstrations. However, obtaining human demonstrations are expensive especially for 6 DoF. In experiments section \ref{sec:Benchmark}, we explicitly compare with these approaches.

In this work, we use a recurrent neural network (RNN) to generate velocity commands. This choice is more efficient over sampling based approaches such as CEM since, RNN can be directly trained using Back Propagation Through Time (BPTT). Furthermore, for sequence prediction tasks RNN is a natural choice over a feed forward neural network. To train our control network, we employ the predictions from the flow network as targets. Thus, our target remains fixed for online trajectory optimization. Note that our linearized predictive model depends only on the depth of the scene, and assuming depth consistency for a smaller horizon, we can write the MPC objective only in terms of control. Hence, our control network could easily be trained in supervised manner by minimising the flow loss:
\begin{equation}
\label{eq:flow_loss}    
\mathcal{L}_{flow} = \lVert \widehat{\mathcal{F}}(\widehat{\mathbf{v}}_{t+1:t+T}) - \mathcal{F}(I_t,I^*)\rVert = \lVert \sum_{k=1}^{T}[L(Z_{t})\widehat{\mathbf{v}}_{t+k}]
 - \mathcal{F}(I_t,I^*)\rVert
\end{equation}
 Where, the RNN $g(\mathbf{v}_t,\theta)$ could be used to generate velocity at next time instance $\widehat{\mathbf{v}}_{t+1} = g(\mathbf{v}_t,\theta)$ given the current velocity.
 Hence, given previous velocity $\mathbf{{v}_{t-1}}$ and tries to improve the predicted velocity and compensate for approximations in predictive model by adjusting its parameters $\mathbf{\theta}$ according such that the flow loss is minimized.
 

\subsection{Control Network}
\label{sec:network_arc}
In this work, we use a Long-Short-Term Memory architecture (LSTM) \cite{lstm} with 5 LSTM cell units as RNN for generating velocities. The sequence length (length of planning horizon as $T$) is a tuning parameter that trades-off between precision and  computation time. In this work we select the sequence length $T =  5$. The network receives previous 6-DoF velocity $\mathbf{v}_{t-1}$ as input and predicts a sequence of $T$ velocities $\widehat{\mathbf{v}}_{t:t+T}$ in 6-DoF. This is multiplied by the interaction matrix $L(z)$ to obtain generated flows $\widehat{\mathcal{F}}_{t:t+T}$. The network is trained online in supervised fashion for $M=100$ iterations by minimizing the flow loss, equation (\ref{eq:flow_loss}), where the target flow is predicted using our flow network. After the network is trained for $M$ iterations we execute the $v_{t+1}$ and observe next image $I_{t+1}$, which is used to predict next target flow $\mathcal{F}(I_{t+1}, I^*)$. Subsequently, the interaction matrix is recomputed $L(Z_{t+1})$ based on updated depth $Z_{t+1}$ using $\mathcal{F}(I_t ,I_{t+1})$ as the proxy. The entire approach can be summarised by algorithm \ref{influx}. 

\begin{algorithm}[t!]
  \caption{Model Predictive Visual Servoing using Online Optimisation}
  \label{influx}
  \begin{algorithmic}[1]
  \Require $I^*$, $\epsilon$ \Comment{Goal Image, convergence threshold}
  \State Initialize $v_{0}$ with random velocity
  \While{$\lVert I - I^* \rVert \leq \epsilon$} \Comment{Convergence criterion}
    \State $I_t := $ get-current-obs() \Comment{Obtain the current RGB observation from sensor}
    \State Predict-Flow ($\mathcal{F}(I_t,I^*)$) \Comment{Predict target flow using flow network}
    \State $L_t := $ compute-interaction-matrix($\mathcal{F}(I_t,I_{t-1})$)
    \For{$m = 0 \to M$} \Comment{Online training the control network}
        \State $\widehat{\mathbf{v}}_{t:t+T} := $ g($v_{t-1}, \theta_m$) \Comment{Velocity predictions from control network }
        \State $\widehat{\mathcal{F}}(\widehat{\mathbf{v}}_{t+1:t+T}) = \sum_{k=1}^{T}[L_t(Z_{t})\widehat{\mathbf{v}}_{t+k}]$ \Comment {Generate flow using prediction model}
        \State $\mathcal{L}_{flow} := \lVert \widehat{\mathcal{F}}(\widehat{\mathbf{v}}_{t+1:t+T}) - \mathcal{F}(I_t,I^*)\rVert.$ \Comment {Compute flow Loss}
        \State $\theta_{m+1}:= \theta_{m}-\eta \nabla \mathcal{L}_{flow}$ \Comment {Update control network parameters}
    \EndFor
    \State $v_{t+1} = \widehat{\mathbf{v}}_{t+1}$ \Comment {Execute the next control command}
  \EndWhile
  \end{algorithmic}
\end{algorithm}\vspace{-0.2cm}

\section{Experiments}\vspace{-0.3cm}
\label{sec:Experiments}

In our present work, we have explained an online training process which enables the robot to learn optimal control commands on the go thus making our controller performance independent of the environment it is deployed in. To validate this, we show both qualitative and quantitative results on photo realistic 3D baseline environments as proposed in \citep{DFVS}. The benchmark comprises of $10$ indoor photo-realistic environments from the Gibson dataset \citep{xia2018gibson} in the Habitat simulation engine \citep{habitat}. We use a free-flying RGB camera as our agent so that agent can navigate in all 6-DoFs without any constraint. To compare our approach we consider following baselines: \textbf{Photometric visual servoing}\cite{photometricvs}: PhotoVS is a learning-free classical visual servoing approach that considers raw image intensities as visual features. \textbf{Servonet}\cite{servonet}, is a supervised deep learning approach that attempts to predict relative camera pose from a given image pair. We benchmark this approach without any retraining/fine-tuning.  \textbf{DFVS}\cite{DFVS}: similar to our approach DFVS employs deep flow representations, however they use classical IBVS controller. To evaluate our controller's performance in isolation from flow network and predictive model we also compare against baselines\textbf{ (a) Fully-connected Neural-Net: } and \textbf{(b) Stochastic Optimization \cite{cem}:} known as the cross-entropy method. We provide more details on the baseline in supplementary section.  

\subsection{Simulation results on Benchmark}
\label{sec:Benchmark}
\begin{figure*}[h!]
\begin{center}
\begin{tabular}{|c|c|c|c|c|c|}
\hline
\begin{tabular}[c]{@{}l@{}}Initial \\ Image\end{tabular} &
\includegraphics[width=18mm, height=12mm]{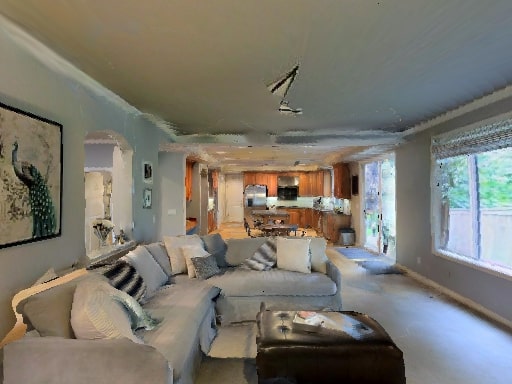} &   
\includegraphics[width=18mm, height=12mm]{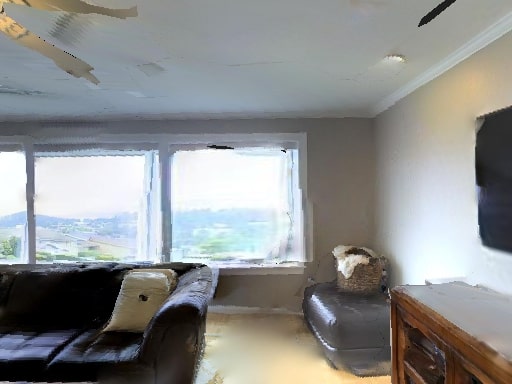} &   
\includegraphics[width=18mm, height=12mm]{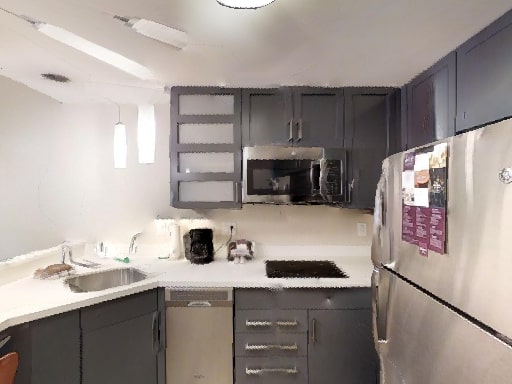} &   
\includegraphics[width=18mm, height=12mm]{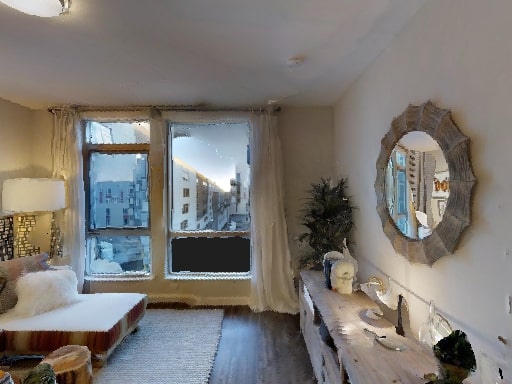} &   
\includegraphics[width=18mm, height=12mm]{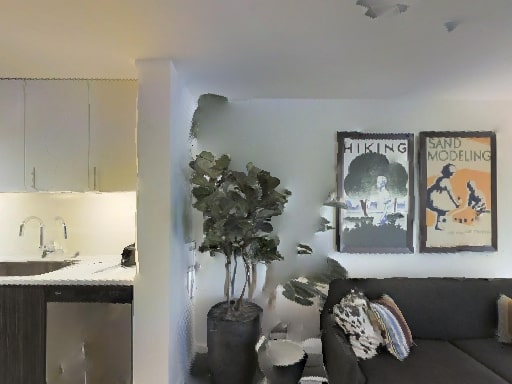}
\\ \hline

\begin{tabular}[c]{@{}l@{}}Desired \\ Image\end{tabular} &
\includegraphics[width=18mm, height=12mm]{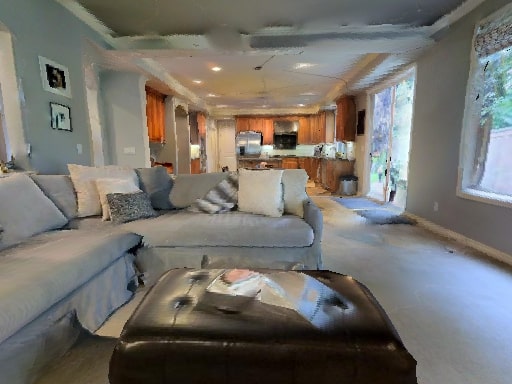} &
\includegraphics[width=18mm, height=12mm]{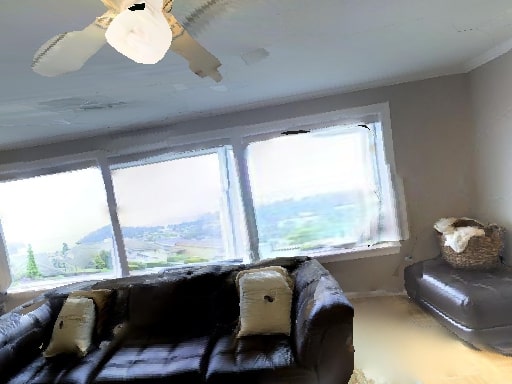} &
\includegraphics[width=18mm, height=12mm]{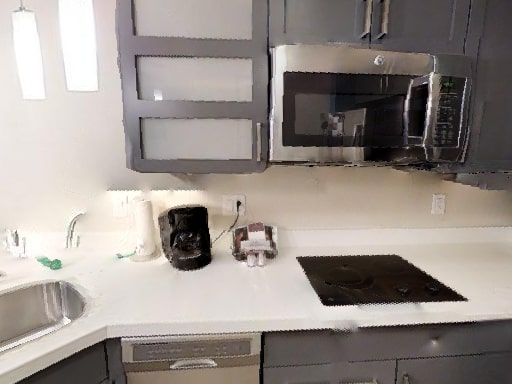} &
\includegraphics[width=18mm, height=12mm]{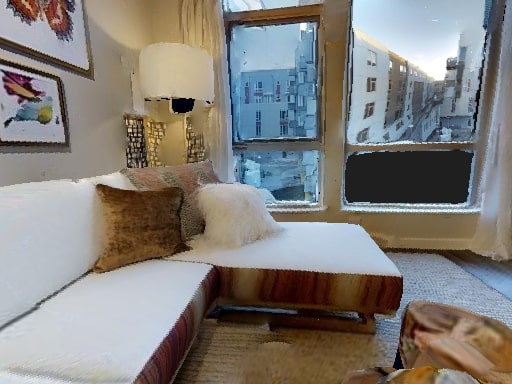} &
\includegraphics[width=18mm, height=12mm]{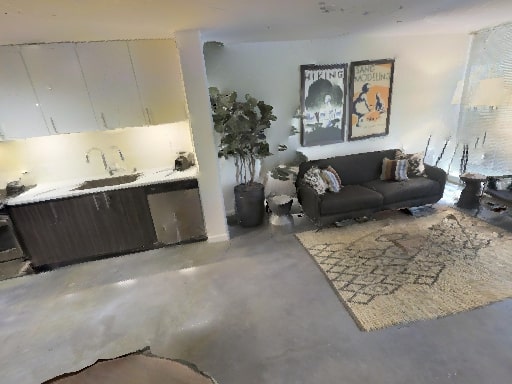} 
\\ \hline

PhotoVS \citep{photometricvs} &
\includegraphics[width=18mm, height=12mm]{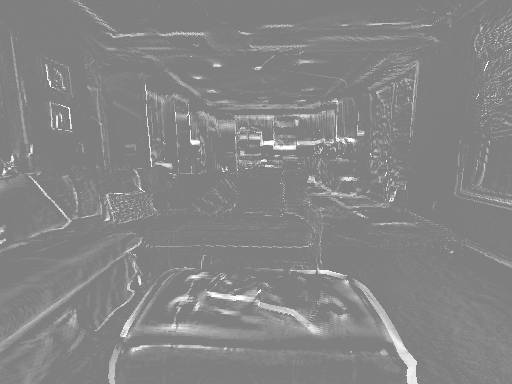} &
\includegraphics[width=18mm, height=12mm]{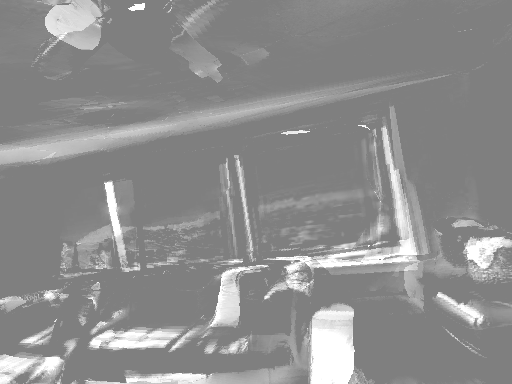} &
\includegraphics[width=18mm, height=12mm]{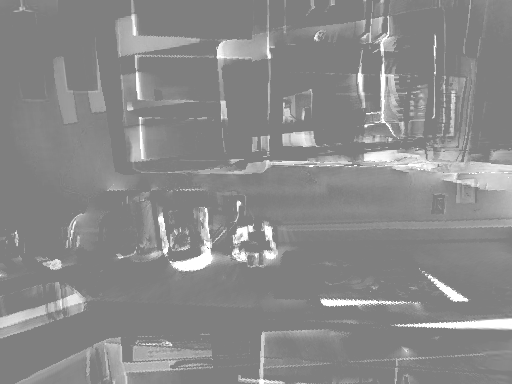} &
\includegraphics[width=18mm, height=12mm]{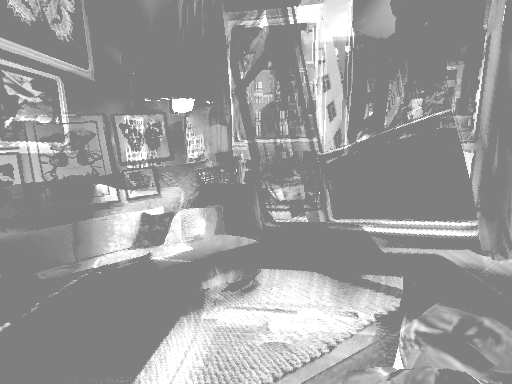} &
\includegraphics[width=18mm, height=12mm]{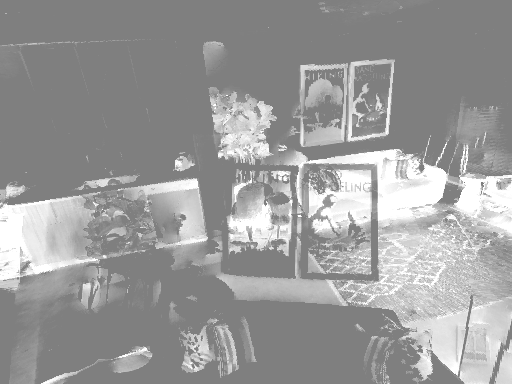} 
\\ \hline

\begin{tabular}[c]{@{}l@{}}Servonet \\ et al.\cite{servonet} \end{tabular}  &
\includegraphics[width=18mm, height=12mm]{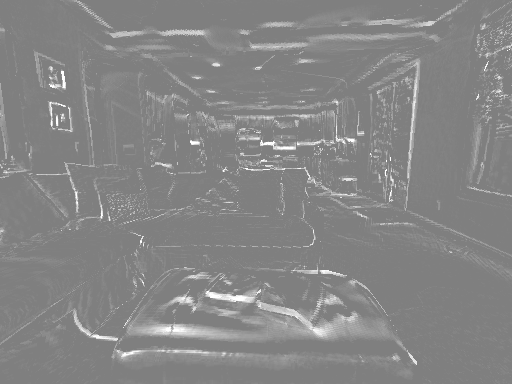} &
\includegraphics[width=18mm, height=12mm]{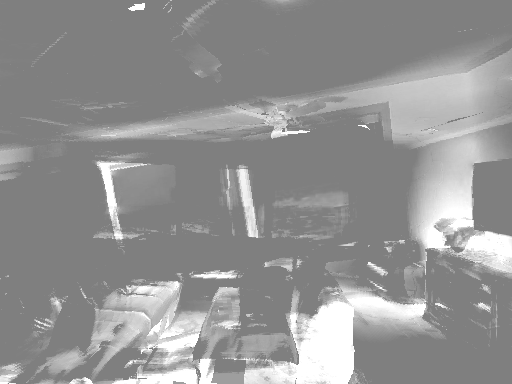} &
\includegraphics[width=18mm, height=12mm]{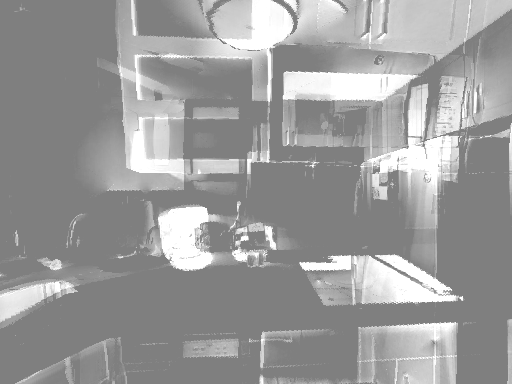} &
\includegraphics[width=18mm, height=12mm]{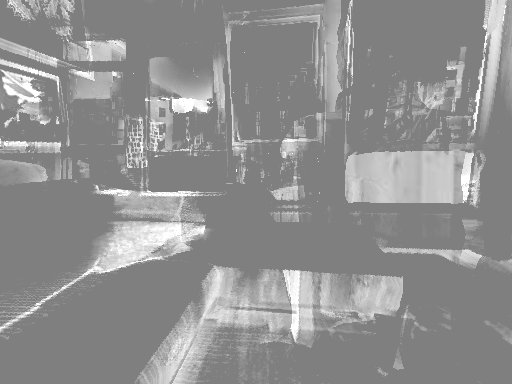} &
\includegraphics[width=18mm, height=12mm]{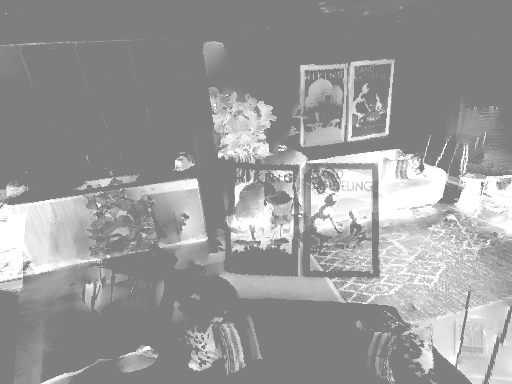} 
\\ \hline

DFVS \cite{DFVS} &
\includegraphics[width=18mm, height=12mm]{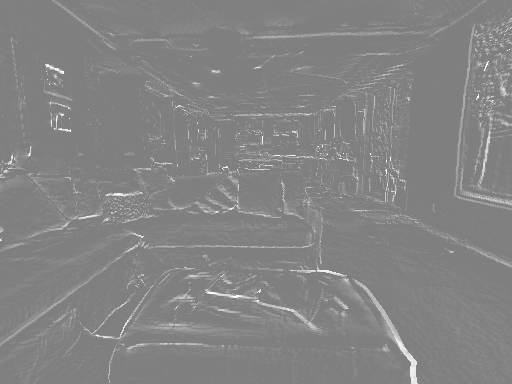} &
\includegraphics[width=18mm, height=12mm]{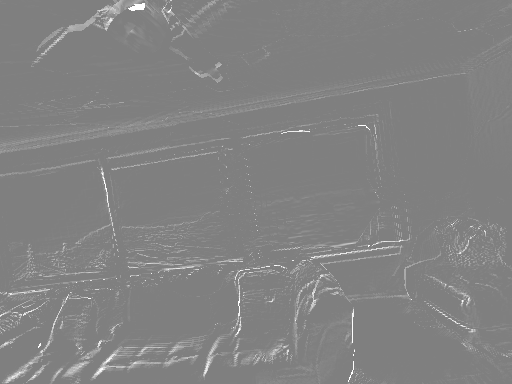} &
\includegraphics[width=18mm, height=12mm]{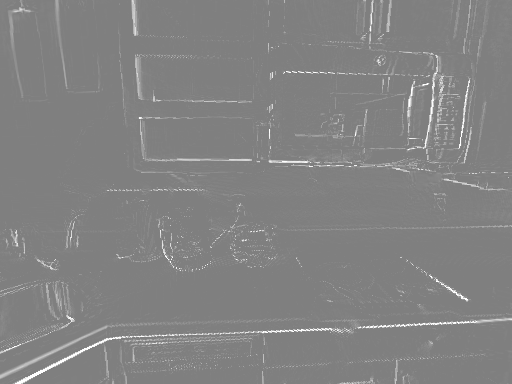} &
\includegraphics[width=18mm, height=12mm]{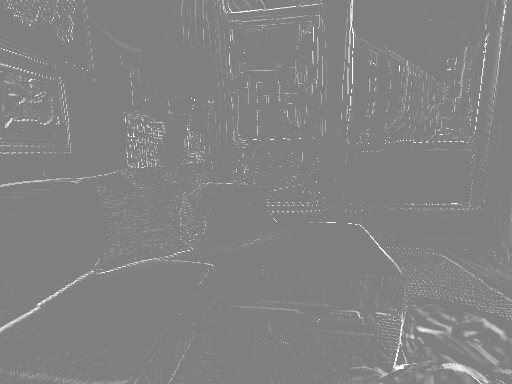} &
\includegraphics[width=18mm,height=12mm]{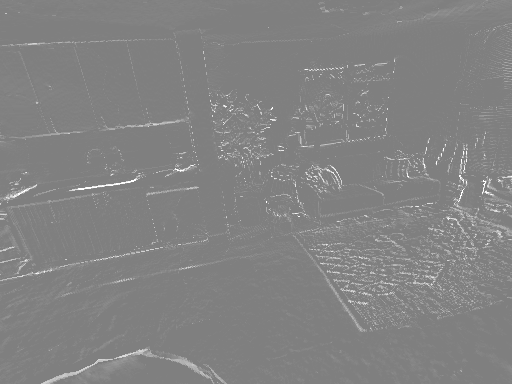} 

\\ \hline

\begin{tabular}[c]{@{}l@{}}MPC +\\ CEM \cite{deepforesight}\end{tabular}  &
\includegraphics[width=18mm, height=12mm]{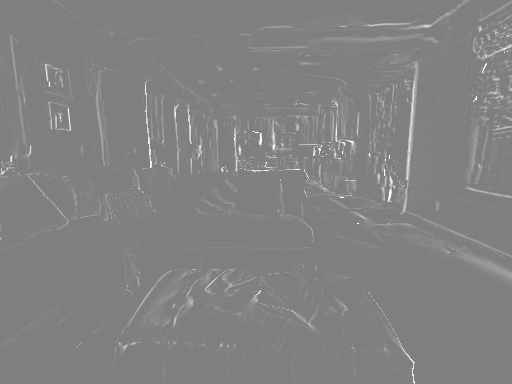} &
\includegraphics[width=18mm, height=12mm]{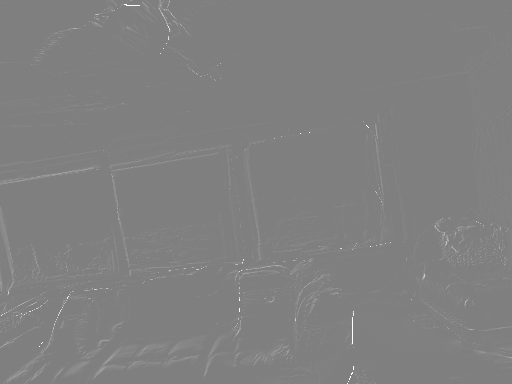} &
\includegraphics[width=18mm, height=12mm]{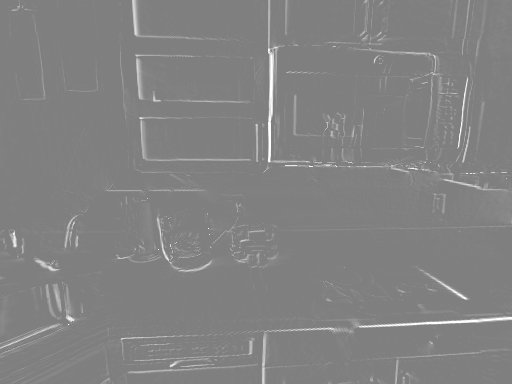} &
\includegraphics[width=18mm, height=12mm]{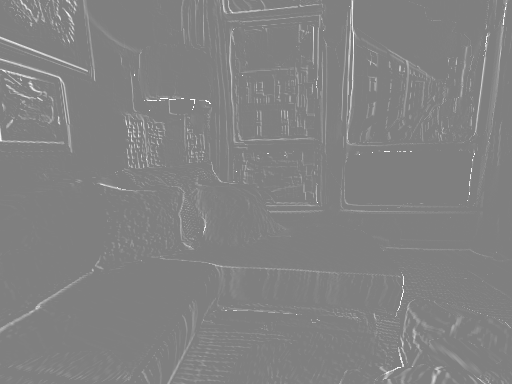} &
\includegraphics[width=18mm, height=12mm]{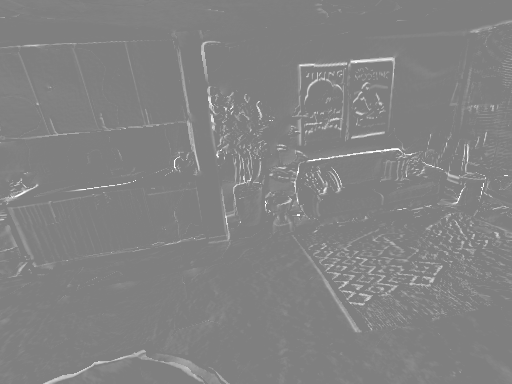} 

\\ \hline

\begin{tabular}[c]{@{}l@{}}MPC +\\ NN {[}Ours{]}\end{tabular} &
\includegraphics[width=18mm, height=12mm]{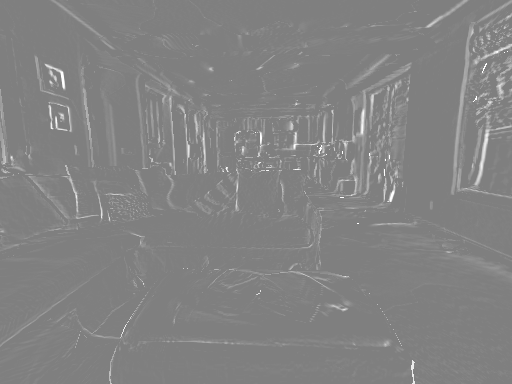} &
\includegraphics[width=18mm, height=12mm]{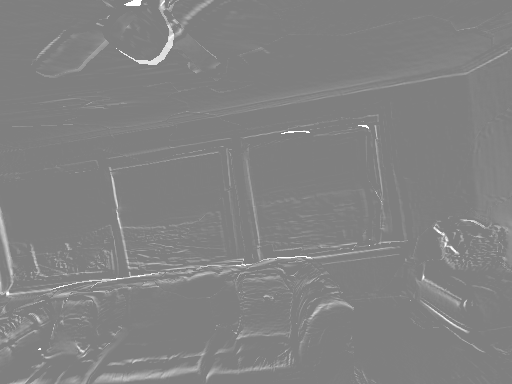} &
\includegraphics[width=18mm, height=12mm]{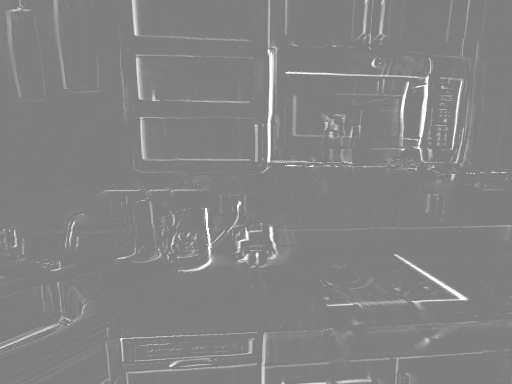} &
\includegraphics[width=18mm, height=12mm]{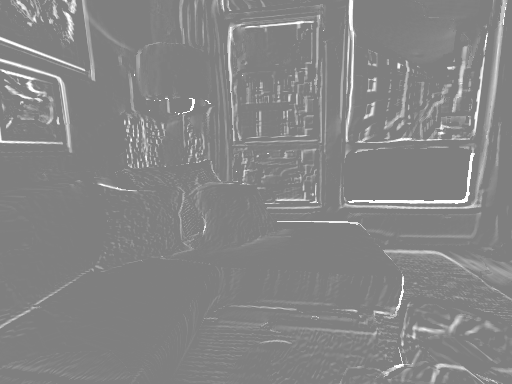} &
\includegraphics[width=18mm, height=12mm]{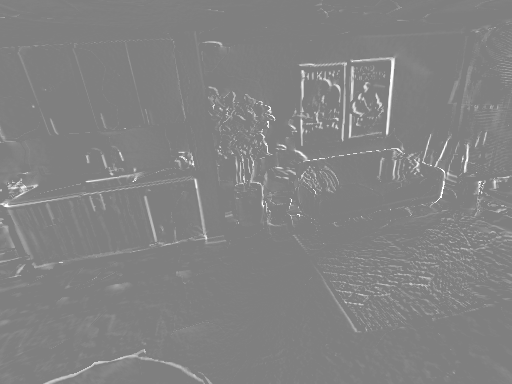} 

\\ \hline

\begin{tabular}[c]{@{}l@{}}MPC +\\ LSTM {[}Ours{]}\end{tabular} &
\includegraphics[width=18mm, height=12mm]{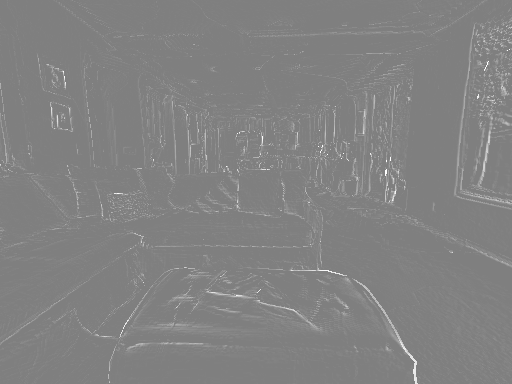} &
\includegraphics[width=18mm, height=12mm]{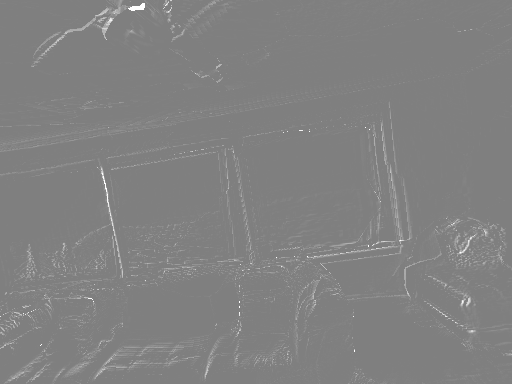} &
\includegraphics[width=18mm, height=12mm]{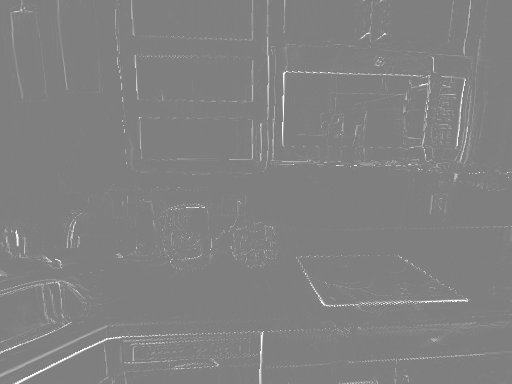} &
\includegraphics[width=18mm, height=12mm]{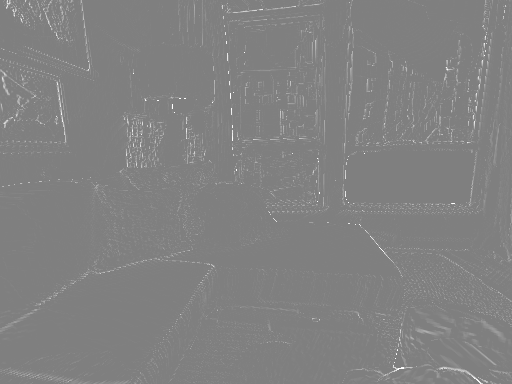} &
\includegraphics[width=18mm, height=12mm]{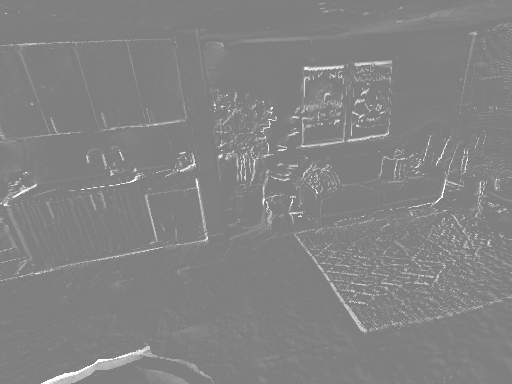} 

\\ \hline


\end{tabular}

\caption{ \textbf{Qualitative Comparison}: We show the photometric error image representation computed between the desired image $I^*$ and the final image on termination for various approaches. It is to be noted that the termination criteria for the run is photometric error of 500 or less. It can be seen that Saxena et al. \cite{servonet} and PhotoVS \citep{photometricvs} don't converge even with large number of iterations hence showing large photometric errors. We compare Flow-depth[ours] with  single-view depth of \citep{DFVS}. Comparisons with MPC + CEM \cite{cem} and MPC + NN [Ours] is also shown as their result justify our choice of LSTM based network. We show other qualitative results of left over baselines in supplementary.}
 \label{fig:bench_qual}
 \end{center}
\vspace{-0.2cm}
 \end{figure*}

We report both qualitative (refer figure \ref{fig:bench_qual}) and quantitative ( refer table \ref{fig:bench_quan}) results on the benchmark. It can be seen from the table \ref{fig:bench_quan} , our controller outperforms the current state-of-the-art Image Based Visual Servoing technique \cite{DFVS} in total number of iterations while achieving marginally superior pose error. PhotoVS \cite{photometricvs} and Servonet \cite{servonet} were unable to converge in 5 and 6 out of 10 scenes in the benchmark respectively. It can also be seen from comparison with the baseline (a) Fully-connected Neural-network that it gets stuck in local minima and results in larger pose error. Furthermore, CEM results in shorter trajectories but takes significantly large amount of time to converge as compared to our approach which is intuitive since sampling approaches are inefficient but statistically complete. We train our control network for 100 iterations in each MPC optimization step and use Adam Optimiser with learning rate of 0.0001. The number of training iterations can be tuned using Early Stopping methods to achieve the best tradeoff between speed and accuracy for realtime control. We use a "Nvidia Geforce GTX 1080-Ti Pascal" GPU to benchmark these approaches.

\begin{table}[h!]\vspace{-0.2cm}
\begin{center}

\begin{tabular}{|l|l|l|l|l
|}

\hline
\multicolumn{1}{|l|}{Approaches} & \begin{tabular}[c]{@{}c@{}}T. Error\\ (meters)\end{tabular} & \begin{tabular}[c]{@{}c@{}}R. Error\\ (degrees)\end{tabular} & \begin{tabular}[c]{@{}c@{}}Tj. Length\\ (meters)\end{tabular} & Iterations 
\\ \hline
\hline
Initial Pose Error & 1.6566         & 21.4166        & -               & -            
\\ \hline
DFVS \citep{DFVS}            & 0.0322         & 1.7088         & 1.7277          & 1009.2222    
\\ \hline
NN + MPC  [ours]         & 0.102              & 2.62            &  $1.16^{*}$             &  $344^{*}$
\\ \hline
CEM \citep{cem} + MPC               & 0.0513         & 0.8644         & \textbf{0.89}          & 859.6667     
\\ \hline
\textbf{LSTM + MPC   [ours]}       & \textbf{0.0296}         & \textbf{0.5977}         & 1.081          & \textbf{557.4444}      
\\ \hline
\end{tabular} 

\caption{\textbf{Quantitative Comparison}:We quantitatively compare the average performance for different approaches across the benchmark scenes proposed in \citep{DFVS} and report following metrics: Initial Pose Error, translation error(T. Error), rotation error(R. Error) in degrees and  trajectory length(Tj. Length) in meters. This table does not include \citep{photometricvs},\citep{servonet}  as they diverge in majority of the environments[5 and 6 respectively of the 10 baseline envs]. }
 \label{fig:bench_quan}
 \end{center}
\vspace{-0.5cm}
 \end{table}
 
\subsection{ Controller Performance and Trajectory Comparison }
\begin{figure}
\begin{center}
\begin{tabular}{ccc}
\includegraphics[width=4.25cm]{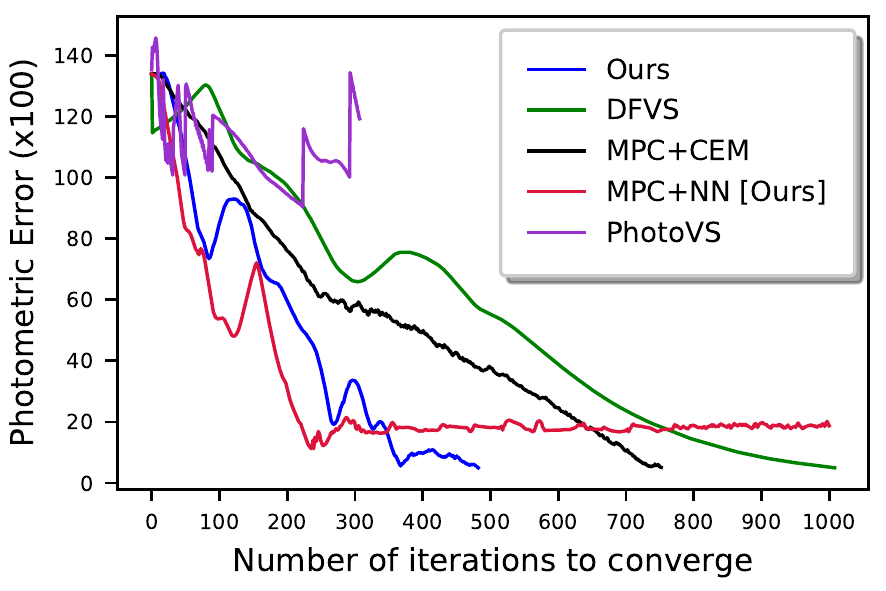} &
\includegraphics[width=4.25cm]{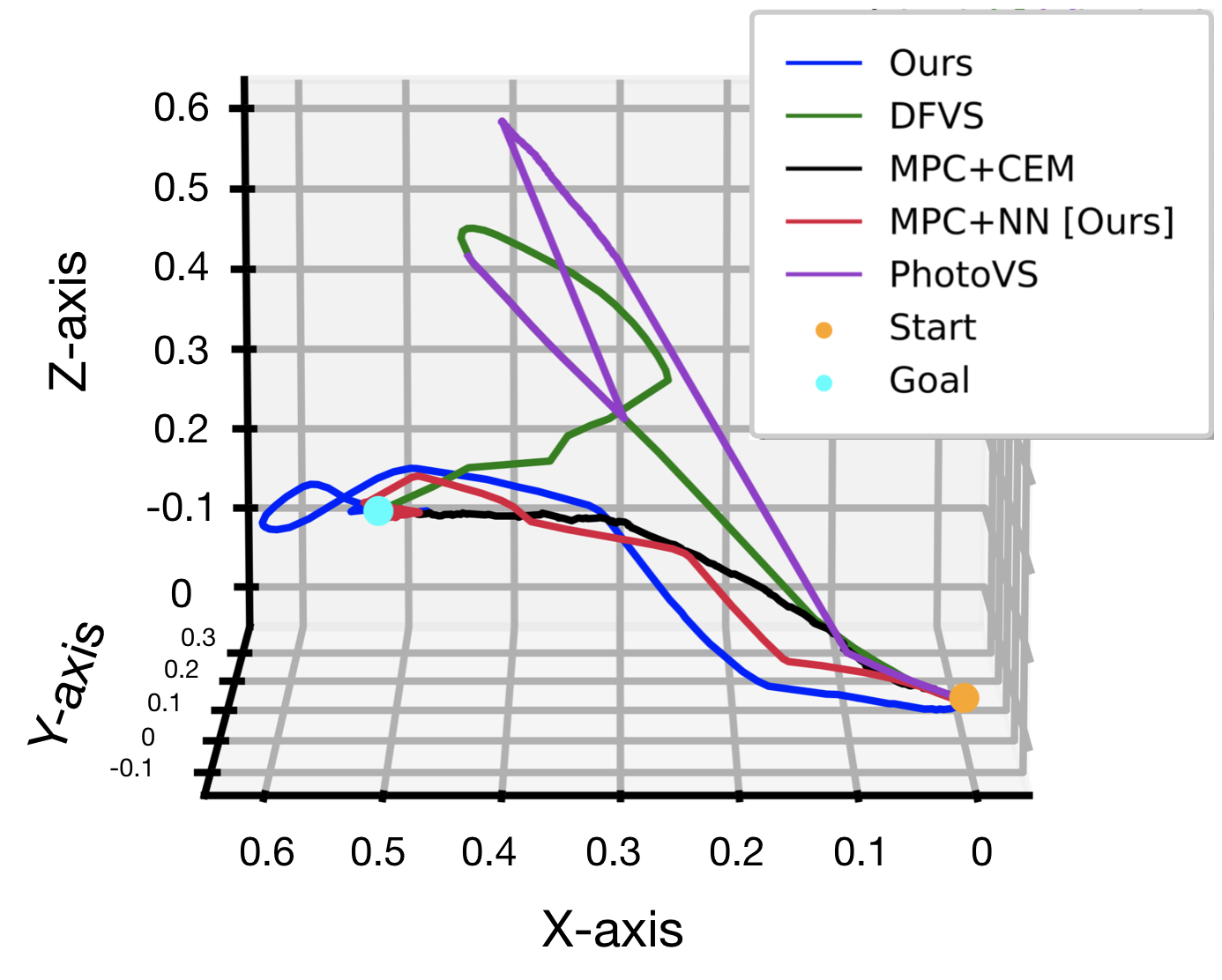} &
\includegraphics[width=4.25cm, height=4cm]{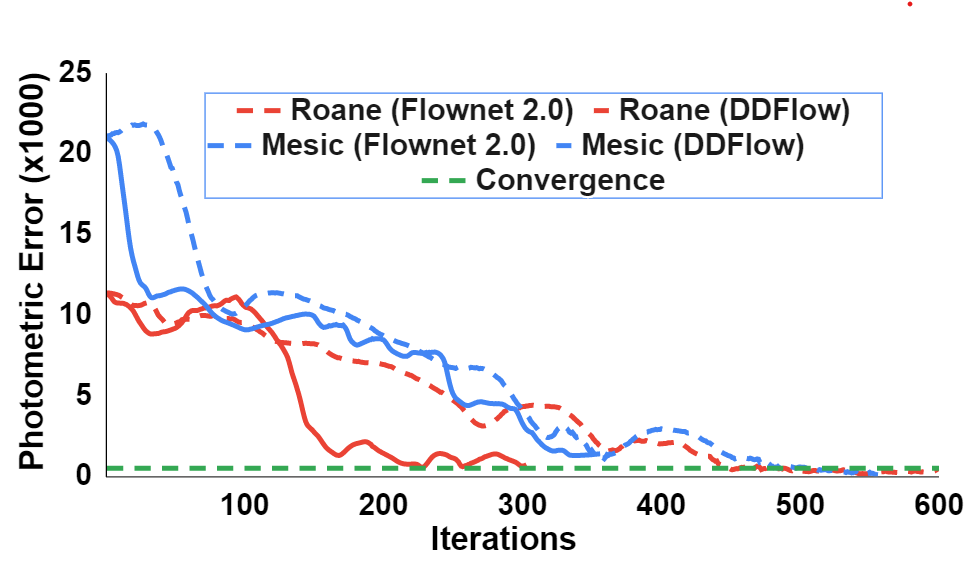} 
\end{tabular}
\caption{\textbf{Controller Performance}: \textbf{Left}: Photometric Error (vs) Number of Iterations - Our method [MPC + LSTM] converges in least number of iterations compared to other approaches. \textbf{Middle}: Camera Trajectory. \textbf{Right}: Controller performance on Unsupervised flow - 
Dashed lines represents Flownet2 \citep{flownet-2} and bold lines represents DDFlow \citep{ddflow} }
\label{fig:controller_supervised}


\end{center}
\end{figure}

\vspace{-0.1cm}
\label{sec:Controller_Trajectory_comp}
We next compare the trajectories taken by other visual servoing methods like \citep{DFVS,photometricvs} and controller performance with a standard feed forward neural network (NN) and CEM \cite{deepforesight} to reach the desired goal. The photometric convergence and trajectory plots are shown in fig. \ref{fig:controller_supervised}. It can be observed from the figure that photometric error steadily reduces to less than $500$. As seen from the trajectory plots, CEM with predictive model tends to reach the goal in an optimal path followed by our approach, while our approach is taking less MPC steps for convergence.


\subsection{Controller performance with Unsupervised flow}\vspace{-0.3cm}
\label{sec:Unsupervised}
To showcase that our deep MPC architecture is also capable of generating optimal control commands even when trained in fully unsupervised manner i.e. here even the flow network is trained without supervision using only images. We use DDFlow \cite{ddflow} instead of Flownet2 \citep{flownet-2} to generate the optical flow. For this experiment, we train our flow network offline on two scenes from the benchmark. For the training details we refer the readers to the supplementary material.  
The photometric error plots in Fig. \ref{fig:controller_supervised} (Right) shows that our controller successfully converges on both the environments, thereby making our pipeline unsupervised. No ground truth or annotations were provided and the flow was learned in an unsupervised fashion without any unlabeled data.  

\subsection{Generalisation to Real-World Noise}\vspace{-0.3cm}
\label{sec:Stability}

In order to demonstrate robustness of our approach we perform tests with simulated noise in the Habitat environment. We add Gaussian noise with mean=0.0 and standard deviation=0.1m in all 6-DoF to control commands. We perform test on environment Mesic from our simulation benchmark \citep{DFVS} and show that our pipeline converges to a photometric error of < 500 in 683 iterations in presence of such a large noise, as compared to 504 iterations without noise. This showcases ability to handle actuation noise and generalize in a real world environment.


%




\section{Conclusion and Future Work}\vspace{-0.4cm}
\label{sec:conclusion}
In this work, we proposed a deep unsupervised model predictive control architecture for visual servoing in 6-DoF. We demonstrated that our approach can generate control commands in continuous state space which can adapt and perform well despite an inaccurate understanding of the underlying system dynamics. Its ability to be trained in an online fashion makes it easy to adapt well in a completely unknown environment.\\
Although the proposed approach addresses some of the most crucial challenges and performs better than state-of-the-art methods proposed on this problem, it suffers due to a few drawbacks. Firstly, this approach assumes that the action space of the robot is unobstructed and there is no chance of collision. Secondly, like all of the visual servoing problems, it requires a small minimum overlap between the current and the desired images which constrains this method to act as a local planner.\\
The above limitations provide an opportunity to extend this approach by incorporating methods to handle collision and integrate this method with a long term global planner. 

\acknowledgments{ This work was done as a collaborative project between International Institute of Information Technology(IIIT), Hyderabad India and  Tata Consultancy Services(TCS), Innovation Labs, India. This work was supported by research grant from TCS Innovation Labs India and  EPSRC UK (project NCNR, National Centre for Nuclear Robotics, EP/R02572X/1) }

\clearpage


\bibliography{example}  
\end{document}